\def\year{2021}\relax
\documentclass[letterpaper]{article} 
\usepackage{arxiv}  
\usepackage{times}  
\usepackage{helvet} 
\usepackage{courier}  
\usepackage[hyphens]{url}  
\usepackage{amsfonts}
\usepackage{graphicx} 
\usepackage{natbib}  
\usepackage{caption} 
\usepackage{amsmath}
\DeclareMathOperator{\sign}{sign}
\usepackage{color}

\title{Adversarial Interaction Attack: Fooling AI to Misinterpret Human Intentions}

\author {

        Nodens Koren,\textsuperscript{\rm 1}
        Qiuhong Ke \textsuperscript{\rm 1}
        Yisen Wang \textsuperscript{\rm 2}
        James Bailey \textsuperscript{\rm 1}
        Xingjun Ma \textsuperscript{\rm 3}
        \thanks{Corresponding author.}
        \\
}
\affiliations {
    \textsuperscript{\rm 1} The University of Melbourne, Victoria, Australia \\ 
    \textsuperscript{\rm 2} Peking University, Beijing, China \\
    \textsuperscript{\rm 3} Deakin University, Geelong, Australia \\
    \{nkoren@student., qiuhong.ke@, baileyj@\}unimelb.edu.au, yisen.wang@pku.edu.cn, daniel.ma@deakin.edu.au
}

\begin{document}

\maketitle

\begin{abstract}
Understanding the actions of both humans and artificial intelligence (AI) agents is important before modern AI systems can be fully integrated into our daily life. In this paper, we show that, despite their current huge success, deep learning based AI systems can be easily fooled by subtle adversarial noise to misinterpret the intention of an action in interaction scenarios. Based on a case study of skeleton-based human interactions, we propose a novel adversarial attack on interactions, and demonstrate how DNN-based interaction models can be tricked to predict the participants' reactions in unexpected ways. From a broader perspective, the scope of our proposed attack method is not confined to problems related to skeleton data but can also be extended to any type of problems involving sequential regressions. Our study highlights potential risks in the interaction loop with AI and humans, which need to be carefully addressed when deploying AI systems in safety-critical applications. 
\end{abstract}

\section{Introduction}
In recent years, Artificial Intelligence (AI) has become much more closely connected to human activity. Many tasks that once used to require human labor, are now gradually being automated and shifted to AI. For instance, 
in order to cope with the COVID-19 pandemic situation, the use of robot workers is being suggested to minimize physical contact between humans. These robot technologies heavily depend on the accuracy of action recognition/prediction and the consequent interaction between humans and machines.

State-of-the-art action recognition and prediction models are deep neural networks (DNNs), due to their capability of modeling complex problems \cite{Si_2019_CVPR, li2019spatio, li2019actional} in an accurate way. Nonetheless, it has also been shown that these models are prone to adversarial examples (or attacks) \cite{biggio2013evasion,szegedy2013intriguing,goodfellow2014explaining}. DNNs can behave erratically when processing inputs with carefully crafted perturbations, even though such perturbations are imperceptible to humans \cite{carlini2017towards,madry2017towards,croce2020reliable,jiang2020imbalanced,wang2020unified}. This has raised security concerns on the deployment of DNN-powered AI systems in security-critical applications such as autonomous driving \cite{eykholt2018robust,duan2020adversarial} and medical diagnosis \cite{finlayson2019adversarial,ma2020understanding}.   Investigating and understanding these abnormalities is a crucial task before machine learning based AI agents can become practical.

In this work, we investigate the adversarial vulnerability of DNN reaction prediction (i.e., regression) models in skeleton-based interactions. Skeleton signals are among one of the most commonly used representations for human or robot motion \cite{zhang2016rgb, wang2018rgb}.  While adversarial attacks have been extensively studied on images \cite{goodfellow2014explaining,su2019one,brown2017adversarial,duan2020adversarial}, very few works have been proposed for skeletons \cite{liu2019adversarial, wang2019smart, zheng2020towards}. 
In comparison to the image space, which is continuous and where pixels can be perturbed freely without raising obvious attack suspicions, the skeleton space is sparse and discrete. It has a temporal nature that needs to be taken into account. Consequently, attacking skeleton-based models requires many more constraints than the image space.

Existing works on attacking skeleton-based models have only considered the single-person scenario, and have all focused towards recognition (i.e., classification) models \cite{liu2019adversarial, wang2019smart, zheng2020towards}. However, interaction scenarios involving two or more characters are essential to the interaction between humans and AI. They should not be overlooked if our ultimate goal is to build AI agents that can fit into our daily life. Neglecting possible attacks might lead to AI agents malfunctioning or behaving aggressively when they are not supposed to.

To close this gap, we propose an Adversarial Interaction Attack (AIA) to test the vulnerability of regression DNNs in skeleton-based interactions involving two characters. Being able to accurately recognize a person's action is important, but it is equally important to be able go a step further and \textit{respond} to the action in an appropriate way. In light of this, the usage of regression models is necessary. We hence modified the output layers of two previous state-of-art models on action recognition. One model was based on a Temporal Convolutional Neural Network (TCN) \cite{BaiTCN2018} and the other was based on Gated Recurrent Units (GRUs) \cite{maghoumi2019deepgru}.  The models were modified to return reactor sequences instead of class labels, and we trained them on skeleton-based interaction data. We examine the performance of AIA attack under both white-box and black-box settings. We show that our AIA attack can easily fool the two regression models to misinterpret the actor's intentions and predict unexpected reactions.  Such reactions have detrimental effects to either the actor or the reactor. Overall, our work reveals potential threats of subtle adversarial attacks on interactions involving AI.

In summary, our contributions are:
\begin{itemize}
\item We propose an adversarial attack approach - Adversarial Interaction Attack (AIA), that is domain-independent, and works for general sequential regression models.

\item We propose an evaluation metric that can be applied to evaluate the performance of sequential regression attacks.  Such a metric is currently missing from the literature. 

\item We empirically show that our AIA attack can generate targeted adversarial action sequences with small perturbations, which fool DNN regression models into making incorrect (possibly dangerous) predictions.

\item We demonstrate via three case studies how our AIA attack may affect human and AI interactions in real scenarios, which motivates the need for effective defense strategies.

\end{itemize}

We highlight that our work is the \textit{first} work on targeted sequential regression attack in a strict manner (i.e. purely numerical outputs without labels of any kind). We do not compare our work to previous works on skeleton-based action recognition as the focus of our work is fundamentally different. Specifically, the goal of our work is to design a new type of attack and evaluation metric that is capable of handling any type of regression-based problems in general. We thus leave the compatibility between our work and the previously proposed anthropomorphic constraints \cite{liu2019adversarial, wang2019smart, zheng2020towards} as a future area of interest.

\section{Related Work}

\subsection{Adversarial Attack}
Adversarial attacks can be either white-box or black-box depending on the attacker's knowledge about the target model. White-box attack has full knowledge about the target model including parameters and training details \cite{goodfellow2014explaining,zheng2019distributionally,croce2020reliable,jiang2020imbalanced}, while black-box attack can only query the target model \cite{chen2017zoo,ilyas2018prior,bhagoji2018practical,dong2019efficient,jiang2019black,bai2020improving} or use a surrogate model \cite{liu2016delving,tramer2017space,dong2018boosting,dong2019evading,andriushchenko2020square,wu2020skip,wang2020unified}. Adversarial attacks can also be targeted or untargeted. Under the classification setting, untargeted attacks aim to fool the model such that its output is different from the correct label, whereas targeted attack aims to fool the model to return a target label of the attacker's interest. White-box attacks can be achieved by solving either the targeted or untargeted adversarial objective using first-order gradient methods \cite{goodfellow2014explaining,kurakin2016adversarial}. Optimization-based methods have also been proposed to achieve the adversarial objective, and at the same time, minimize the perturbation size \cite{carlini2017towards,chen2017ead}. 

Most of the above existing attacks were proposed for images and classification models, and the perturbation is usually constrained to be small (eg. $\|\epsilon\|_{\infty}=8$ for pixel value in $[0,255]$) so as to be imperceptible to human observers. Defenses against adversarial attacks have also been explored on image dataset \cite{madry2017towards,zhang2019theoretically,wang2019convergence,bai2019hilbert,wang2020improving,wu2020adversarial,bai2021improving}.

\noindent\textbf{Attacking Regression Models.}
Untargeted regression attacks can be derived from classification attacks by simply attacking the regression loss \cite{8846746}. 
However, it is more difficult to perform targeted regression attack such that the model outputs a target sequence. This is because, unlike classification models that contain a finite set of discrete labels, regression models can have infinitely many possible outcomes. Hence, most existing attacks on regression models have focused on the untargeted setting.
\citet{10.1007/978-3-030-36708-4_39} proposed a univariate regression loss with the goal of changing the outputs of EEG-based BCI regression models to a value that is at least $t$ away from the natural outcome. This loss function guarantees only that the adversarial output will be at a specified distance away from the natural output. It does not constrain how large or small the output can actually become. Compared to \citet{Cheng2020Seq2SickET}, the order of target sequences is more significant for our problem.
In natural language processing (NLP), \citet{Cheng2020Seq2SickET} proposed a targeted attack towards recurrent language models. This work aims to replace arbitrary words in the output sequence with a small set of target adversarial keywords, regardless of their order and occurrence position. While word embedding can be used to evaluate attack performance on language models, an appropriate performance metric is still lacking in the field of interaction prediction, making it difficult to evaluate the effectiveness of an attack.

None of the existing works have implemented an attack that is able to change the whole output sequence completely. In our work, we propose such an attack, which can change the entire output sequence with target frames appearing in our desired order.

\noindent\textbf{Adversarial Attack on Action Recognition.}
Previous attacks on skeleton-based action recognition have proposed several constraints based on extensive study of anthropomorphism and motion. These include postural constraints as the maximum changes in joint angles, and inter-frame constraints based on the notion of velocities, accelerations, and jerks \cite{liu2019adversarial, wang2019smart, zheng2020towards}. Additionally, \citet{liu2019adversarial} utilized a Generative Adversarial Network (GAN) loss to model anthropomorphic plausibility.
These constraints are distinct from our work, but could potentially be employed in combination with our proposed attack to improve naturalness of adversarial action sequences.

\subsection{Interaction Recognition and Prediction}
The use of skeleton data has gained its popularity in action recognition and prediction research. Owing to the fact that reliable skeleton data can be easily extracted from modern RGB-D sensors or RGB camera images, these techniques can be easily extended to practical applications \cite{kiwon_hau3d12}. 
One benchmark interaction dataset is the SBU Kinect Interaction Dataset. Different from most skeleton-based action recognition datasets that focus on studying single-person activities, the SBU Kinect Interaction Dataset captures various activities with two characters involved. Predicting interactions is a much harder task in comparison to predicting single-person activities, due to the complexity and the non-periodicity of the problem \cite{kiwon_hau3d12}. Specifically, in the interaction scenario, two characters are involved. However, the contribution from each character may not be equal. For instance, interactions such as approaching and departing have only one active character; another character remains steady over all time frames. 

Convolutional Neural Networks (CNNs) \cite{du2015skeleton,nunez2018convolutional,li2017skeleton} and Recurrent Neural Networks (RNNs) \cite{du2015hierarchical} are two popular choices to tackle the interaction recognition problem. Models from the RNN family such as Gated Recurrent Unit (GRU) and Long Short-Term Memory (LSTM) are commonly chosen for interaction recognition, because it is natural for them to handle sequential data. \citet{maghoumi2019deepgru} proposed a recurrent-based model namely DeepGRU, that was able to reach state-of-the-art performance. Temporal Convolutional Networks (TCNs) are also a common choice of model when dealing with spatio-temporal data. TCNs, just like RNNs, can take sequences of any length. TCNs rely on a causal convolution operation to ensure no information leakage from future to the past \cite{BaiTCN2018}. TCN is also a previous state-of-art model \cite{kim2017interpretable} and a component adopted by many latest works on skeleton-based action recognition \cite{meng2018human, yan2018spatial}.

In this paper, we will modify the DeepGRU network proposed by \citet{maghoumi2019deepgru} and the TCN network proposed by \citet{BaiTCN2018} for interaction prediction and examine their vulnerability to our proposed attack based on the SBU Kinect Interaction Dataset.

\section{Proposed Adversarial Interaction Attack}
In this section, we first provide a mathematical formulation of the targeted adversarial sequence attack problem. We then introduce the loss functions used by our AIA attack.

\noindent\textbf{Overview.} Intuitively, the goal of our AIA attack is to deceive the \emph{reactor} AI agent into thinking that the \emph{actor} is doing a different specific action by making minor changes to the positions of the \emph{actor's} joints or the angles between joints.
 The reactor agent will consequently respond by performing the reaction that is targeted by the attack.

\subsection{Formal Problem Definition}
A skeleton sequence with $T$ frames can be represented mathematically as the vector $\mathbf{X} = (\mathbf{x}_1, \mathbf{x}_2, ..., \mathbf{x}_T)$ where $\mathbf{x}_i$ is a skeleton representation of the $i^{th}$ frame, which is a vector consists of  3D-coordinates of the human skeleton joints. More specifically,  $\mathbf{x}_i \in \mathbb{R}^{N \times 3}$, where N denotes the number of the joints. In our approach, we flattened $\mathbf{x}_i$ into $\mathbb{R}^{3N}$.

First, we define the formal notion of interaction. Suppose the two characters in a two-person interaction scenario are \emph{actor A} and \emph{reactor B}. The task of an interaction  prediction model $f$ is to predict an appropriate reaction (i.e., skeleton) $\mathbf{y}_t$ at each time step $t$ for reactor B based on the observed skeleton sequence of actor A $(\mathbf{x}_1, \cdots, \mathbf{x}_{t})$. This can be written mathematically as:
$$f(\mathbf{x}_1, \cdots, \mathbf{x}_{t-1},\mathbf{x}_{t}) = \mathbf{y}_t.$$

Given an input skeleton sequence
$\mathbf{X} = (\mathbf{x}_1, \mathbf{x}_2, ..., \mathbf{x}_T)$, an adversarial target skeleton sequence $\mathbf{Y}'= (\mathbf{y}'_1, \mathbf{y}'_2, ..., \mathbf{y}'_T)$, and a prediction model $f: R^{T \times 3N} \rightarrow R^{T \times 3N}$, the goal of our AIA attack is to find an adversarial input sequence $\mathbf{X}'=(\mathbf{x}'_1, \cdots, \mathbf{x}'_{T})$ by solving the following optimization problem:

\begin{equation}\label{eq:obj}
\begin{aligned}
& \min_{\mathbf{X}'} \sum_{t \in T} \|\mathbf{x}_t' - \mathbf{x}_t \|_{\infty} \\
& s.t. \;\; \sum\limits_{t \in T} \|f(\mathbf{x}'_1, \cdots ,\mathbf{x}'_t) - \mathbf{y}_t' \|_2 < \kappa,
\end{aligned}
\end{equation}
where, $\|\cdot\|_{p}$ is the $L_{p}$ norm, and $\kappa \geq 0$ is a \emph{tolerance factor}, which serves as a cutoff that distinguishes whether the output sequence is recognizable as the target reaction. This gives us more flexibility when crafting the adversarial input sequence $\mathbf{X}'$ because the acceptable target sequence is non-singular; the output sequence does not need to be exactly the same as the target sequence to resemble a particular action. We empirically determine this factor based on informal user survey in Section \ref{sec:5.1}.
Intuitively, the above objective is to find a sequence $\mathbf{X}'$ with minimum perturbation from $\mathbf{X}$, such that the distance between the output and the target is less than $\kappa / T$ on average for each time step.

\subsection{Adversarial Loss Function}
Our goal is to develop a mechanism that crafts an adversarial input sequence which solves the above optimization problem given any target output sequence, while also maintaining the naturalness of the adversarial input sequence.
In order to achieve this goal, we propose the following adversarial loss function:
\begin{equation}\label{eq:adv}
    \mathcal{L}_{adv} = \mathcal{L}_{spatial} + \lambda \mathcal{L}_{temporal},
\end{equation}
where the $\mathcal{L}_{spatial}$ loss term minimizes the spatial distance between the output sequence and the target sequence, and the $\mathcal{L}_{temporal}$ loss term maximizes the coherence of the perturbed input sequence so as to maintain the naturalness of the adversarial input sequence. 

\noindent\textbf{Spatial Loss.}
The spatial loss term aims to generate adversarial output sequences that are visually similar to the target reaction sequences; that is, its objective is to minimize the spatial distance between the output joint locations and the \textit{neighbourhood} of the target joints for every time step. Following the formulation of the relaxed optimization problem in \eqref{eq:obj}, we use the $L_2$ norm to measure the distance between two sets of joint locations:

\begin{equation}\label{eq:spatial}
    \mathcal{L}_{spatial} = \sum\limits_{t \in T}
    \inf\{\|f(\mathbf{x}'_1, \cdots ,\mathbf{x}'_t) - \mathbf{p}_t\|_2 \; | \; \mathbf{p}_t \in S_t\}
\end{equation}
with $S_t$ being an $(N \mbox{-} 1)$-sphere defined by:
\begin{equation}\label{eq:spatial2}
S_t(\mathbf{y}'_t, \eta) = \{\mathbf{p}_t \in \mathbb{R}^{3N} \;|\; \|\mathbf{p}_t - \mathbf{y}'_t\|_2 = \eta\}.
\end{equation}
Here, $\eta = \kappa / T$ is the mean of the enabling tolerance factor $\kappa$ in equation \eqref{eq:obj} over time $T$.

\noindent\textbf{Temporal Loss.}
The temporal loss term is to guarantee the naturalness of the generated adversarial input sequence. Specifically, the movement of each joint should be continuous in time, and motions with abrupt huge change or teleportation should be penalised.  The $\mathcal{L}_{temporal}$ term achieves this goal by maximizing the coherence of each element in the perturbed input sequence with respect to its neighboring elements in the temporal dimension. This gives:
\begin{equation}\label{eq:temporal}
    \mathcal{L}_{temporal} = \sum\limits_{t \in T} (\|\mathbf{x}'_t - \mathbf{x'}_{t-1}\|_2
    + \|\mathbf{x}'_t - \mathbf{x'}_{t+1}\|_2)
\end{equation}
Note that a scaling factor $0 \leq \lambda \leq 1$ is introduced in front of $\mathcal{L}_{temporal}$ to balance the two loss terms.

We use the first-order method Project Gradient Descent (PGD) \cite{madry2017towards} to minimize the combined adversarial loss iteratively as follows:
\begin{equation}
\begin{aligned}
    &\mathbf{X}'_0 = \mathbf{X} \\
    &\mathbf{X}'_{m+1} = \Pi_{\mathbf{X}, \epsilon} \big(\mathbf{X}'_m - \alpha \cdot \sign (\nabla_{\mathbf{X}'_m} \mathcal{L}_{adv}(\mathbf{X}'_m, \mathbf{Y'}))\big)
\end{aligned}
\end{equation}
where, $\Pi_{\mathbf{X}, \epsilon}(\cdot)$ is the projection operation that clips the perturbation back to $\epsilon$-distance away from $\mathbf{X}$ when it goes beyond, $\nabla_{\mathbf{X}'_m}\mathcal{L}_{adv}(\mathbf{X}'_m, \mathbf{Y'})$ is the gradient of the adversarial loss to the input sequence, $m$ is the current perturbation step for a total number of $M$ steps, $\alpha$ is the step size and $\epsilon$ is the maximum perturbation factor. The sequence $\mathbf{Y'}$ for a target reaction can be either customized or sampled from the original dataset.

\begin{figure*}[h!]
    \includegraphics[width=17.6cm]{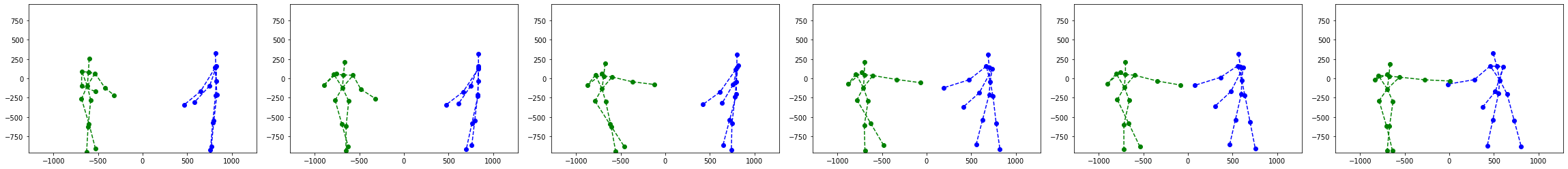}
    \includegraphics[width=17.6cm]{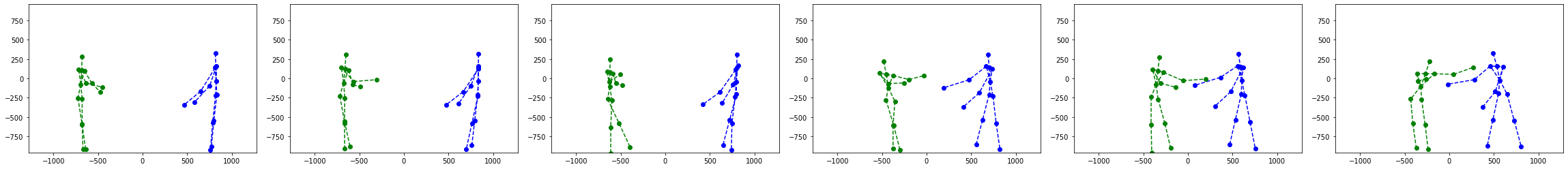}
    \caption{Side-by-side comparison of Case Study 1 ‘handshaking’ to ‘punching’. Top-Bottom: original prediction, adversarial prediction. Blue character: input, green character: output.}
    \label{fig:1}
\end{figure*}

\begin{figure*}[h!]
    \includegraphics[width=17.6cm]{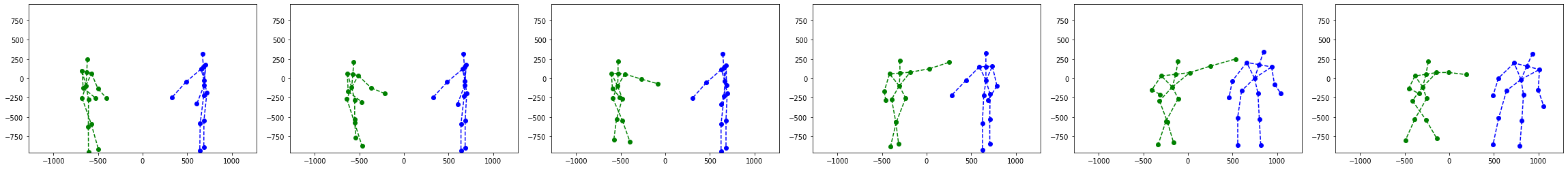}
    \includegraphics[width=17.6cm]{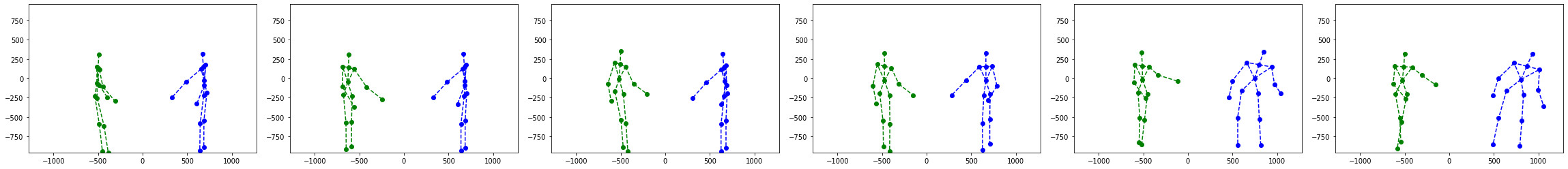}
    \caption{Side-by-side comparison of Case Study 2 ‘punching’ to ‘handshaking’. Top-Bottom: original prediction, adversarial prediction. Blue character: input, green character: output.}
    \label{fig:2}
\end{figure*}

\begin{figure*}[h!]
    \includegraphics[width=17.6cm]{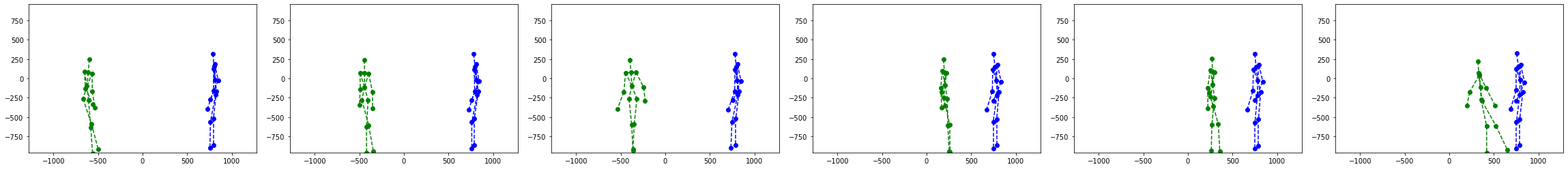}
    \includegraphics[width=17.6cm]{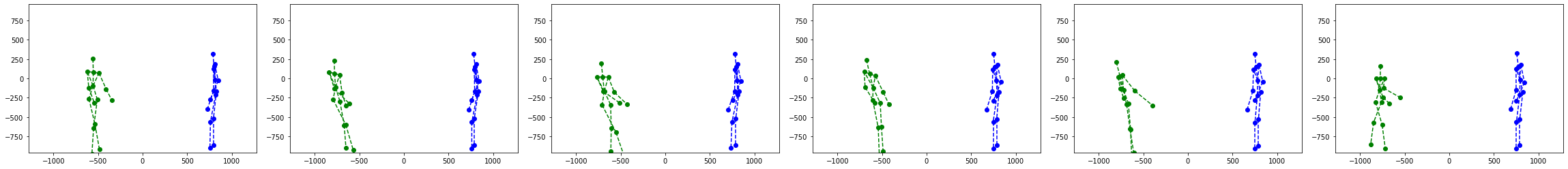}
    \caption{Side-by-side comparison of Case Study 3 ‘approaching’ to ‘remaining’. Top-Bottom: original prediction, adversarial prediction. Blue character: input, green character: output.}
    \label{fig:3}
\vspace{-0.1in}
\end{figure*}

\section{Overview on Several Case Studies}
In this section, we conduct case studies on three selected sets of attack objectives that can be easily associated with real scenarios and can serve as  motivations behind our approach. Detailed experimental settings can be found in Section \ref{sec:5}. The dynamic versions of the case studies and more examples are provided in the supplementary materials.

\subsection{Case Study 1: `handshaking' to `punching'}
Figure \ref{fig:1} illustrates a successful AIA attack that fools the model to predict a `punching' action for the reactor (the green character) as a response to the adversarially perturbed `handshaking' action of the actor (the blue character). Note that the perturbation only slightly changed the actor's action.
This reveals an important safety risk that needs to be carefully addressed before machine learning based AI agents can be widely used in human daily life. 
Suppose that we are at an AI interactive exhibition, a participant would like to shake hands with an AI robot agent. He gradually extends his hand, sending out an interaction request to the AI agent and is expecting the AI agent to respond to his handshaking invitation by shaking hand with him. However, instead of reaching its hands out gently, the AI agent decided to punch the participant in the face because the participant's body does not stay straight. It would be extremely hazardous if the human character unintentionally wiggled his body in a pattern similar to the adversarial perturbation introduced in this case study. While the actual chance of this happening is extremely low due to the high complexity of data in both the spatial and the temporal dimensions, this threat might nevertheless happen if AI workers become widely deployed worldwide. In this case, the human is a victim by inadvertently performing an adversarial attack (wiggling their body).

\subsection{Case Study 2: `punching' to `handshaking'}
In this case study we consider a case opposite to the previous one, where human exploiters are capable of attacking AI agents actively and derive benefit from being active attackers.
In the future, it could become a common practice to utilize AI agents to complete dangerous tasks so as to lower the chance of human operators incurring injuries or fatalities. Security guard is one such  job that might be taken over by an AI agent. Imagine a secret agency that hires AI security guards is invaded by intruders and is placed in a scenario where combat becomes necessary. The AI guard will  fail in its  role if the invaders know how to apply effective adversarial attacks towards it.  This is the case in Figure \ref{fig:2} where the model was fooled to suggest `handshaking' for the reactor (the green character) rather than `punching'.

\subsection{Case Study 3: `approaching' to `remaining'}
Finally, Case Study 3 demonstrated in Figure \ref{fig:3} examines the case of how a cheater might be able to bypass an AI agent's detection.
Whilst automatic ticket checkers have been widely adopted, manual ticket checking is still required for numerous situations. For instance, public transportation companies may want to check whether a passenger has paid for the upgrade fee if he or she is in a first class seat. Now suppose that 
a public transportation company decides to hire AI agents to do the ticket checking job. The public transportation company will lose a huge amount of income if passengers know how to stop the ticket checkers from `approaching' as in Figure \ref{fig:3}, or even change their `approaching' response to `departing'.

\section{Empirical Understanding of AIA}\label{sec:5}

\subsection{Tolerance Factor $\kappa$}\label{sec:5.1}
The objective of AIA attack is defined with respect to a tolerance factor $\kappa$ (see \eqref{eq:obj}, \eqref{eq:spatial} and \eqref{eq:spatial2}), which is a flexible metric that distinguishes whether the output sequence is close to the targeted adversarial reaction.
Because there are many factors involved, such as the character's height, handedness, and the direction the character is facing, conventional distance metrics such as $L_1$ and $L_2$ norms are not suitable to define precisely what the pattern of a specific action should look like. Therefore, we determine the value of $\kappa$ based on human perception via an informal user survey. 

In order to obtain appropriate values for $\kappa$ to evaluate whether an attack is successful, we randomly sampled 5 out of 8 sets of attack objectives and presented them to 82 human judges, including computer science faculties and students. Each objective set is composed of an action-reaction pair and contains output sequences generated from 6 different values of $\epsilon$ (from left to right in ascending order). For each sample set, we asked the human judges to choose the leftmost sequence they believe is performing the target reaction. Sampled objectives and the responses from the 82 human judges are recorded in Table \ref{tab:1}.

Based on the responses from the 82 human judges, we computed the tolerance factor $\kappa$ in the optimization problem defined in  \eqref{eq:obj} based on the average of
\begin{equation}\label{eq:kappa}
    \sum\limits_{t \in T} \|f(\mathbf{x}'_1, \cdots ,\mathbf{x}'_t) - \mathbf{y}_t' \|_2
\end{equation}
over the 5 sample objective sets. The calculation of \eqref{eq:kappa} for each objective set is based on the minimum $\epsilon$ polled from the 82 human judges, and the corresponding value of $\kappa$ is then selected as the optimal value (boldfaced in Table \ref{tab:1}).

Note that, $\kappa$ serves as a topological boundary between the natural and the adversarial \textit{outputs}, whereas $\epsilon$ is a maximum perturbation constraint that we don't want the \textit{input} perturbation to go beyond.

\begin{table*}[!htbp]
\caption{Responses from the 82 human judges. The optimal $\kappa$ for each attack objective is highlighted in \textbf{bold}. }
\label{tab:1}
\centering
\small
\begin{tabular}{llllllll}
\hline
$\epsilon =$ & 0.075 & 0.15 & 0.225 & 0.3 & 0.375 & 0.45 \\
\hline
Handshaking    & 1 ($\kappa = 90.9$) & 4 ($\kappa = 84.28$) & \textbf{44 ($\mathbf{\kappa = 79.52}$)} & 3 ($\kappa = 74.49$) & 12 ($\kappa = 45.04$)  & 14 ($\kappa = 35.03$)  \\
Punching    & \textbf{58 ($\mathbf{\kappa = 52.04}$)} & 13 ($\kappa = 47.63$) & 6 ($\kappa = 43.97$) & 3 ($\kappa = 41.76$) & 0 ($\kappa = 39.14$)  & 2 ($\kappa = 34.91$)  \\
Kicking    & 3 ($\kappa = 100.61$) & \textbf{71 ($\mathbf{\kappa = 93.17}$)} & 7 ($\kappa = 86.57$) & 1 ($\kappa = 80.68$) & 0 ($\kappa = 47.47$) & 0 ($\kappa = 35.36$) \\
Departing    & 0 ($\kappa = 85.03$) & 7 ($\kappa = 76.78$) & \textbf{26 ($\mathbf{\kappa = 71.77}$)} & 12 ($\kappa = 67.58$) & 1 ($\kappa = 41.78$) & 10 ($\kappa = 32.70$) \\
Pushing    & 6 ($\kappa = 28.66$) & 3 ($\kappa = 26.55$) & 2 ($\kappa = 25.16$) & 14 ($\kappa = 23.98$) & \textbf{49 ($\mathbf{\kappa = 22.77}$)} & 5 ($\kappa = 21.31$) \\
\hline
\end{tabular}
\end{table*}

\begin{figure*}[!ht]
    \includegraphics[width=17.6cm]{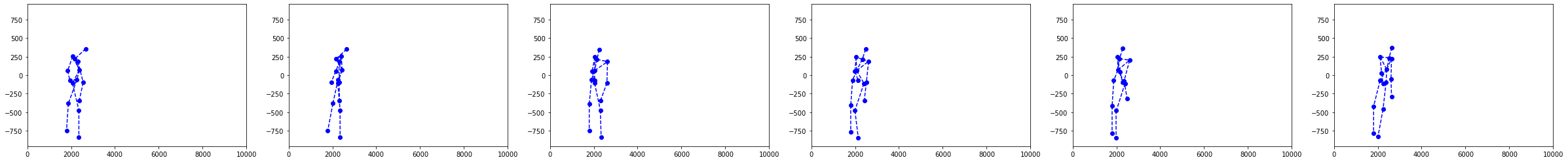}
    \includegraphics[width=17.6cm]{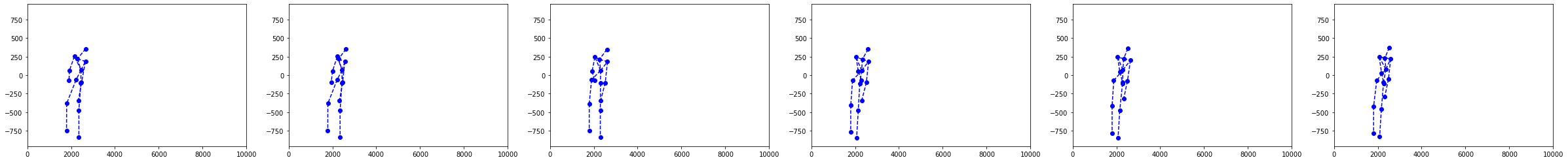}
    \caption{Adversarial input action sequences generated by our AIA attack with (bottom row, and $\lambda = 0.1$) or without (top row) the temporal constraint $\mathcal{L}_{temporal}$.}
    \label{fig:5}
\vspace{-0.1in}
\end{figure*}

\subsection{Effect of the Temporal Constraint}
Here, we study the effect of the temporal constraint $\mathcal{L}_{temporal}$ defined in  \eqref{eq:temporal} on the naturalness of the generated adversarial input action sequence.
Specifically, we investigate how the input skeleton sequence changes in the depth axis as that is the only perturbed dimension throughout our experiments. Our hypothesis is that this additional factor will enable our AIA attack to find adversarial input sequences that change more smoothly with respect to time.

We demonstrate visually a comparison between adversarial sequences generated with and without the temporal constraint in Figure \ref{fig:5}. The top sequence is an adversarial input sequence generated with the $\mathcal{L}_{temporal}$ term removed, whereas the bottom sequence is an adversarial input sequence generated with $\lambda=0.1$ scaling factor applied to the $\mathcal{L}_{temporal}$ term. In comparison to the previous experiment, we plot the skeletons from the depth-y point of view as we are more interested in visualizing the perturbation.

As shown in Figure \ref{fig:5}, it is observable that in general, the top sequence has more abrupt changes in body position between each time step. This almost never happens in the bottom sequence. More specifically, in the bottom sequence, when a larger change to the body posture is necessary, the change is always preceded by smaller changes in the same direction. In contrast, in the top sequence, any large changes can take place in just one time step. This type of aggressive change should be avoided as much as possible, as it could make the attack more easily detectable.

\section{Performance Evaluation}
We conduct two sets of experiments to evaluate the effectiveness (white-box attack success rate) and the transferability (black-box attack success rate) of our AIA attack.

\subsection{Experimental Settings}\label{sec:6.1}
\noindent\textbf{Dataset.}
We conduct our experiments on the benchmark SBU Kinect Interaction Dataset, which is composed of interactions of eight different categories, namely `approaching', `departing', `kicking', `punching', `pushing', `hugging, `handshaking', and `exchanging'. It contains 21 sets of data sampled from 7 participants using a Microsoft Kinect sensor, with approximately 300 interactions in total. Each character's information is encoded into 15 joints with the $x$, $y$, and depth dimensions. The values of $x$ and $y$ fall within $[0, 1]$, and depth in $[0, 7.8125]$.

In order to extract action sequences and reaction sequences, we partitioned each interaction into two individual sequences corresponding to each character respectively. One sequence will be used as the action (input) and another will be used as the reaction (output). Due to the lack of data in this dataset, we trained our response predictors from the perspectives of both characters. With this belief, we used the skeleton sequences of both characters as input data independently. That is, for each interaction sequence $\mathbf{x} = \mathbf{x}_1 \frown \mathbf{x}_2$, we create two input/target pairs $(\mathbf{x}_1, \mathbf{x}_2)$ and $(\mathbf{x}_2, \mathbf{x}_1)$.

\noindent\textbf{Models and Training.}
We adopted one convolutional model, TCN \cite{BaiTCN2018}, and one recurrent model, DeepGRU \cite{maghoumi2019deepgru}, and modified them such that the models predict sequences instead of categorical labels. 
Our TCN model has 10 hidden layers with 256 units in each layer and our DeepGRU model follows \citet{maghoumi2019deepgru} exactly with the output being a linear layer instead of the attention-classifier framework.
We trained each model on the preprocessed dataset for 1,000 epochs using the Adam optimizer with a learning rate of 0.001. We held out sets s01s02, s03s04, s05s02, s06s04 in the original dataset as our test set.

\noindent\textbf{Attack Setting.}
In all experiments, we used the same step size of $\alpha=0.03$ and run our AIA attack for $M=400$ iterations. In addition, we used the Adam optimizer with a learning rate of $1e-3$ to maximize the adversarial loss function $\mathcal{L}_{adv}$. The scaling factor $\lambda$ for the temporal loss term $\mathcal{L}_{temporal}$ was set to 0.1. The tolerance factor $\kappa$ was selected for each target reaction based on our previous informal user survey in Section \ref{sec:5.1} (the exact values can be found in Table \ref{tab:1}).

\subsection{Effectiveness of our AIA Attack}\label{sec:6.2}
In this experiment, we examine the effectiveness of our AIA attack under the white-box setting with different values of maximum perturbation $\epsilon$ allowed.
In order for an attack to be considered successful, it has to satisfy two conditions: 1) the adversarial output sequences need to be recognizable as the target reaction (related to $\kappa$), and 2) the adversarial input sequences need to be visually similar enough compared to the natural input sequences such that it can circumvent security detection (related to $\epsilon$). Hence, the smaller the $\epsilon$ the attack can work under, the more effective the attack is.

Without loss of generality and in order to control the overall change to the input sequence, we perturbed only the depth dimension for each joint. This makes it much easier to visualize perturbations. On a side note, this is a stricter optimization problem with constraints in comparison to the original proposed problem. The outcome of this experiment is thus applicable to the original problem as well.
\subsubsection{Adversarial Targets.}
We created 8 sets of target reactions, corresponding to all 8 interactions in the SBU Kinect Interaction Dataset. The objective of each set of targets is to change the output reactions of all test data into one specific target reaction. We then perform targeted adversarial attacks based on these objectives over a range of $\epsilon$ values.

\begin{figure}[!ht]
    \includegraphics[width=0.49\linewidth]{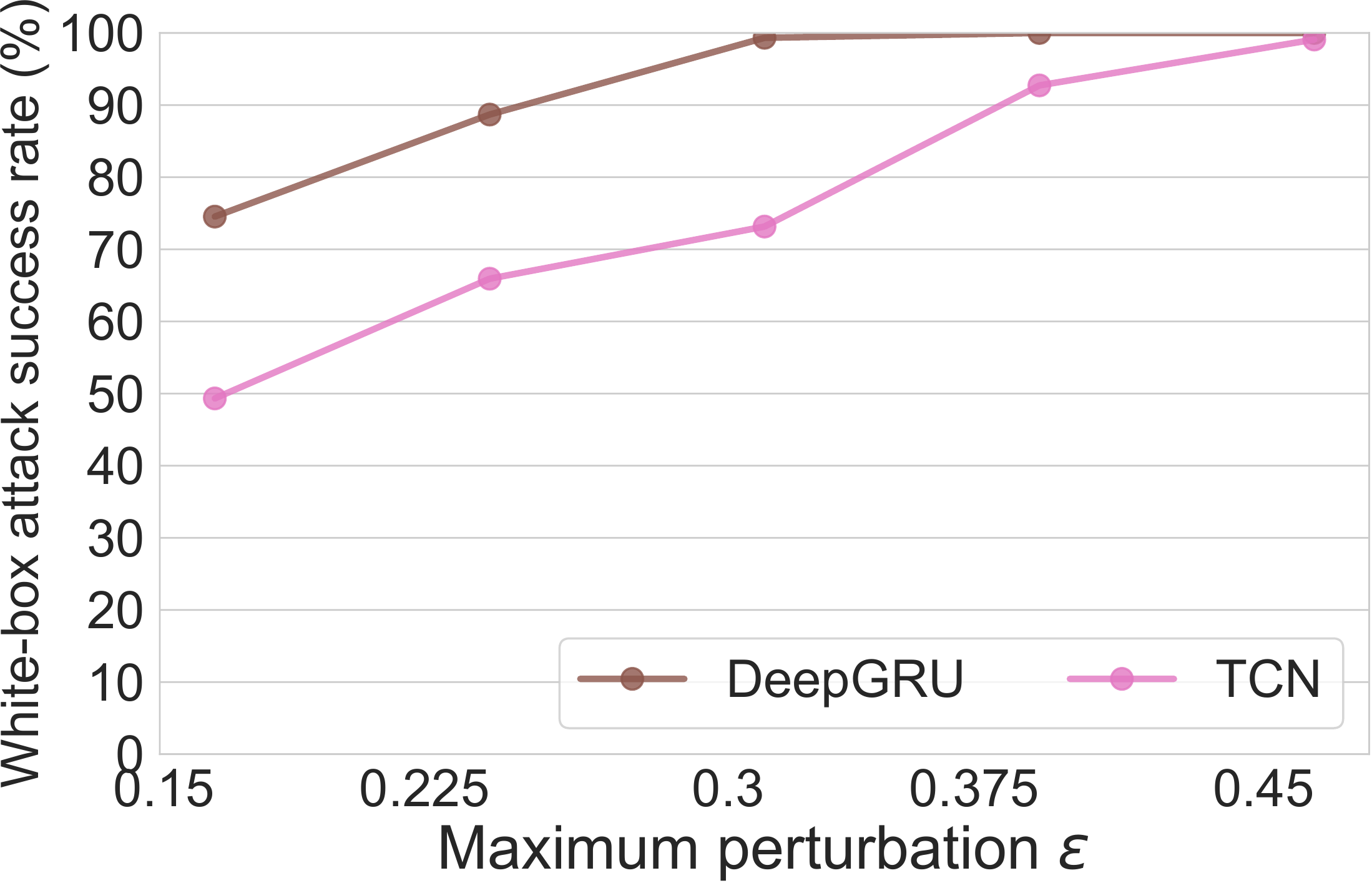}
    \includegraphics[width=0.49\linewidth]{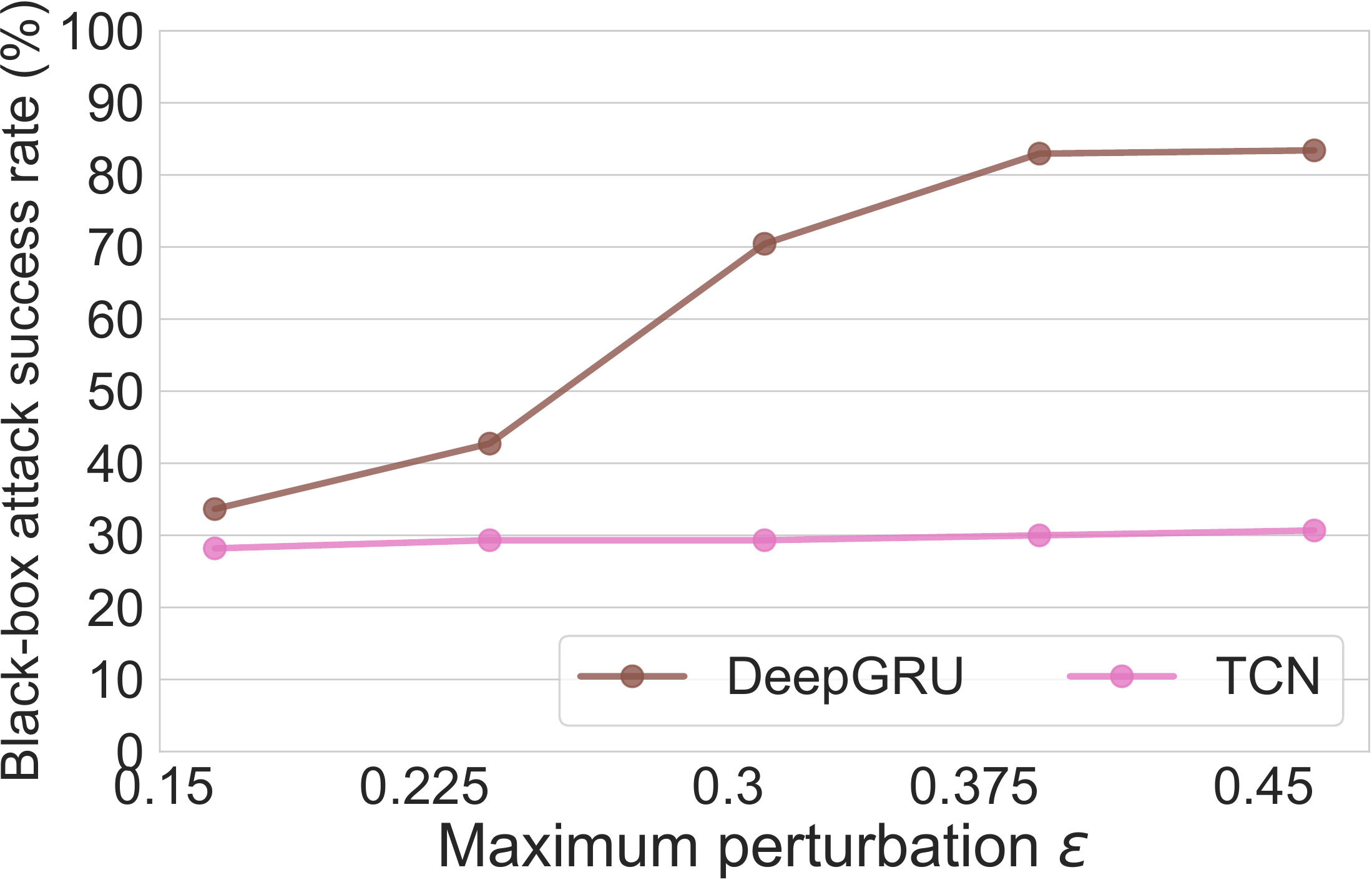}
    \caption{Average white-box (left) and black-box (right) attack success rate of our AIA attack on TCN and DeepGRU.}
    \label{fig:6}
\vspace{-0.1in}
\end{figure}

We consider an attack to be successful if the sum term in \eqref{eq:obj} computed on the test datum is less than the human-determined $\kappa$ based on the sample sets. Otherwise we consider the attack to have failed. The average attack success rates over all 8 target sets under various $\epsilon$ are reported for both models in the left subfigure of Figure \ref{fig:6}. We used the $\kappa$ sampled from human judges to evaluate attack success rates for objectives 1 to 5. We expect the $\kappa$ to be generalizable to unseen reactions, so we used the average $\kappa$ over 5 objective sets to evaluate the remaining 3 attack objectives.

\subsubsection{Results.}
On average, with a perturbation factor $\epsilon$ of 0.225 to 0.3, our AIA attack is able to alter almost all output sequences of the DeepGRU model into any target sequence. On the other hand, a larger $\epsilon$ of 0.375 to 0.45 is necessary for AIA to achieve a similar level of performance on the TCN model. In general, the TCN model is more robust to our attack than the DeepGRU model. However, under this white-box setting, we were able to achieve a 100\% attack success rate on almost all target sets for both models.

We close up this experiment with a conclusion that when model parameters are available, our AIA attack is very effective towards deep sequential regression models. Note that the depth value falls within [0, 7.8125]. This indicates that our AIA algorithm is able to accomplish most attack objectives with small perturbations of 2\% to 5\% to natural input sequences. More generally, our attack works for any target sequences, not only confined to specific interactions from the dataset. This enables our attack method to work for both targeted and untargeted goals. For untargeted goals, the attacker simply needs to pick an arbitrary target sequence that is far enough from the original sequence.

\subsection{Black-Box Transferability}
In addition to white-box effectiveness, we examine how transferable our attack is. An adversarial example generated based on one model is said to be transferable if it can also fool other independently trained models. In this experiment, we examine robustness of the TCN model and the DeepGRU model towards adversarial examples generated based on each other.

\subsubsection{Black-box Setting.}
We employed the same metric established in Section 5.1 to determine an attack to be successful or not. To evaluate how strong our attack is under the black-box setting, we reused the adversarial input sequences in the previous experiment. We feed all adversarial sequences generated based on one model into another and inspect their effectiveness when used to attack unseen model. In other words, we use adversarial sequences generated based on the DeepGRU model into the TCN model and vice versa. The average black-box attack success rates over a range of $\epsilon$ are reported for both models in the right subfigure of Figure \ref{fig:6}.

\subsubsection{Results.}
Surprisingly, adversarial examples generated from the TCN model are remarkably strong. With an $\epsilon$ value of 0.375 to 0.45, adversarial actions generated from the TCN model successfully fooled the DeepGRU model more than 80\% of the time for almost all attack objectives.
Along with the results in Section \ref{sec:6.2}, this substantiates that our AIA attack has high transferability in addition to being effective.

We also observed that adversarial actions generated from the DeepGRU model are rather weak on the TCN model under the black box setting. It is only able to achieve an average success rate of 30\% irrespective to the maximum perturbation $\epsilon$ permitted. The TCN model is more robust than DeepGRU in the white-box setting.
We suspect that this is because the convolutional layers used in TCN are more robust than the gated recurrent units of DeepGRU. 
Specifically, in order to fool the TCN model, the attack needs to take into account the high level feature maps between the convolutional layers. 
However, adversarial examples generated from the DeepGRU model might not be able to fool the convolutional layers of TCN because these high level features were not taken into consideration in the first place. Note that, while being relatively more robust, TCN also leads to more transferable attacks.
We leave further inspection to this disparity as a future work.

\section{Conclusion}
In this paper, we presented a framework for attacking general spatio-temporal regression models. We proposed the first targeted sequential regression attack that is capable of altering the entire output sequence completely - Adversarial Interaction Attack (AIA). On top of that, we also defined an evaluation metric that can be adopted to evaluate the performance of adversarial attacks on sequential regression problems. We demonstrated on variants of two previous state-of-art action recognition models, TCN and DeepGRU, that our AIA attack is very effective. Additionally, we showed that our AIA attacks are highly transferable if referenced from proper models. We also discussed through three case studies, how AIA might impact interactions between human and AI in real scenarios.  We hope this serves to motivate careful consideration about how to effectively incorporate  AI based agents into human daily life.

\bibliographystyle{aaai}
\bibliography{ref}

\end{document}